\documentclass[journal]{IEEEtran}

\ifCLASSINFOpdf
\else
\fi

\usepackage{url}
\usepackage{booktabs}
\usepackage{algorithm}
\usepackage{multirow}
\usepackage{graphicx}
\usepackage{xcolor}
\usepackage[table]{xcolor}
\usepackage{makecell}
\usepackage{subcaption}
\usepackage{enumitem}
\usepackage{threeparttable}
\usepackage{balance}
\usepackage{amsmath}

\definecolor{deltapos}{HTML}{1A7F37}   %
\definecolor{deltaneg}{HTML}{C1272D}   %

\definecolor{mpcol}{RGB}{226,240,237}  %
\definecolor{macol}{RGB}{244,234,226}  %

\usepackage{cleveref}
\crefname{figure}{Fig.}{Figs.}
\crefname{table}{Tab.}{Tabs.}
\crefname{equation}{Eq.}{Eqs.}
\crefname{section}{Sec.}{Secs.}
\crefname{algorithm}{Alg.}{Algs.}
\Crefname{figure}{Figure}{Figures}
\Crefname{table}{Table}{Tables}
\Crefname{equation}{Equation}{Equations}
\Crefname{section}{Section}{Sections}
\Crefname{algorithm}{Algorithm}{Algorithms}
\usepackage[caption=false,font=footnotesize]{subfig}
\definecolor{yellowgreen}{HTML}{9ACD32}
\definecolor{purple}{HTML}{8A2BE2}   
\definecolor{red}{HTML}{FF0000}     
\definecolor{morphblue}{HTML}{1E90FF} 

\usepackage{xcolor}
\definecolor{best}{HTML}{2ca02c}
\definecolor{worst}{HTML}{d62728}

\begin{document}
\title{Rank-1 Identity Consensus Predicts \\ Gallery Enrollment in 1:N Face Matching \\ More Accurately than Score Thresholding}
\author{
\IEEEauthorblockN{Gabriella Pangelinan\IEEEauthorrefmark{1}, 
Aman Bhatta\IEEEauthorrefmark{2}, 
Michael C. King\IEEEauthorrefmark{1}, and 
Kevin W. Bowyer\IEEEauthorrefmark{2}}\\
\IEEEauthorblockA{\IEEEauthorrefmark{1}Florida Institute of Technology, Melbourne, FL}
\IEEEauthorblockA{\IEEEauthorrefmark{2}University of Notre Dame, Notre Dame, IN}
}
\maketitle

\begin{abstract}
In operational 1:N face identification, a crucial question arises for each probe: is this person enrolled in the gallery or not? The stakes are high and asymmetric. Reject a mate-present (MP) probe and a valid lead is lost; accept a mate-absent (MA) probe and every returned candidate is a false identification, at worst a wrongful arrest. Approaches in the literature predominantly threshold match scores, but scores shift substantially with image quality and gallery size and composition, making thresholds fixed before deployment brittle under realistic operational conditions. Our prior work introduced 1-consistency, the only method based on rank consensus across multiple independently trained matchers. Probes are labeled MP if all matchers return the same rank-1 identity.

This work stress-tests 1-consistency across 36 (gallery, probe quality) scenarios spanning four quality levels and two structural axes: images per identity and total enrolled identities. We benchmark against two score-thresholding methods that bracket what any deployed threshold could achieve—one realistic, one best-case but operationally infeasible. Fixed Score-Thresholding (FST), calibrated once on baseline conditions, collapses asymmetrically as quality degrades: MP recall falls below 2\% while MA recall holds near 100\%. Oracle Score-Thresholding (OST), re-tuned per scenario, is the best any threshold could theoretically do—yet for degraded probes 1-consistency matches it, with zero tuning. The two differ mainly in error type (OST favors MP recall, 1-consistency favors MA recall), but on one axis 1-consistency does not merely match the oracle: when it labels a probe MP, it returns the correct mate 97–100\% of the time versus OST's 66–84\% under severe degradation. 

The takeaway is that 1-consistency gives the accuracy of the oracle without the impossible requirement: it sets no threshold, so it needs no advance knowledge of the conditions a probe will arrive in, which is what makes it usable.
\end{abstract}

\IEEEpeerreviewmaketitle

\section{Introduction}
\label{sec:introduction}

In operational contexts, 1:N face identification systems contend with
two persistent challenges. The first is image quality. Probes often
come from surveillance footage, with poor resolution, blur, occlusion,
and other quality problems. Quality has a drastic effect on
identification performance: state-of-the-art systems with near-perfect
accuracy on high-quality images~\cite{FRVT_Identification} exhibit
FPIRs upwards of 20\% and substantial demographic disparities when
probe and/or gallery quality is poor~\cite{bhatta2024impact,
pangelinan2024analyzing, pangelinan2026dual}.

The second challenge is uncertainty about gallery enrollment. An
investigator searching surveillance footage cannot know in advance
whether a subject of interest has ever been enrolled.
This open-set identification problem---determining whether a probe's
true identity is present in the gallery---is fundamental to
investigative workflows. If a probe is mate-absent (MA), any returned
candidate is by definition a false positive, which can lead to wasted
investigative effort and, in the worst case, a wrongful arrest.
Conversely, if a mate-present (MP) probe is misclassified as MA, valid
investigative leads are missed.

Existing approaches to classifying gallery membership rely
predominantly on similarity scores, either through fixed or adaptive
thresholds~\cite{frvt-demographic, guenther2017toward} or through
probabilistic modeling of score tail
distributions~\cite{scheirer2011metarecognition, scheirer2013toward,
scheirer2014probability}. However, score distributions shift substantially across operational variables like image quality and gallery structure, and vary across demographic groups ~\cite{bhatta2024impact, pangelinan2024analyzing, pangelinan2026dual}. A fixed threshold cannot accommodate changing operational variables, and would likely be unfair across demographics. An adaptive threshold would require recalibration for every combination of probe quality, gallery composition, and demographic makeup—infeasible in deployment, since the combination a probe will present is not known in advance.

\begin{figure}[t]
\centering
\includegraphics[width=.9\columnwidth]{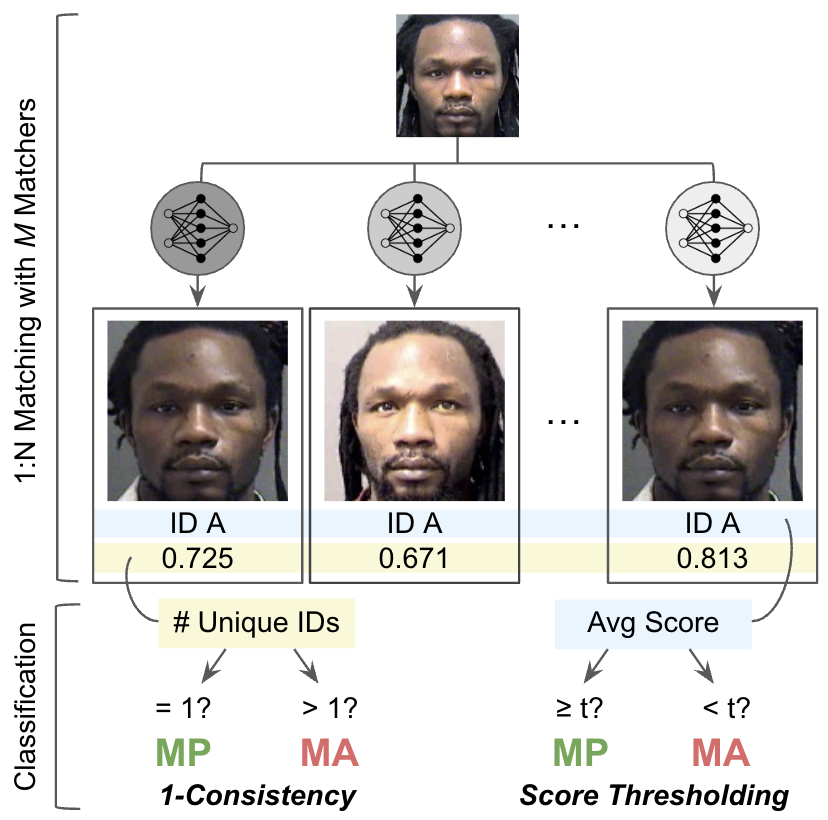}
\caption{Visualization of 1-consistency versus score thresholding classification using $M$ independently trained matchers.}
\label{fig:title_fig}
\end{figure}

\begin{figure*}[t]
\centering
\includegraphics[width=.83\textwidth]{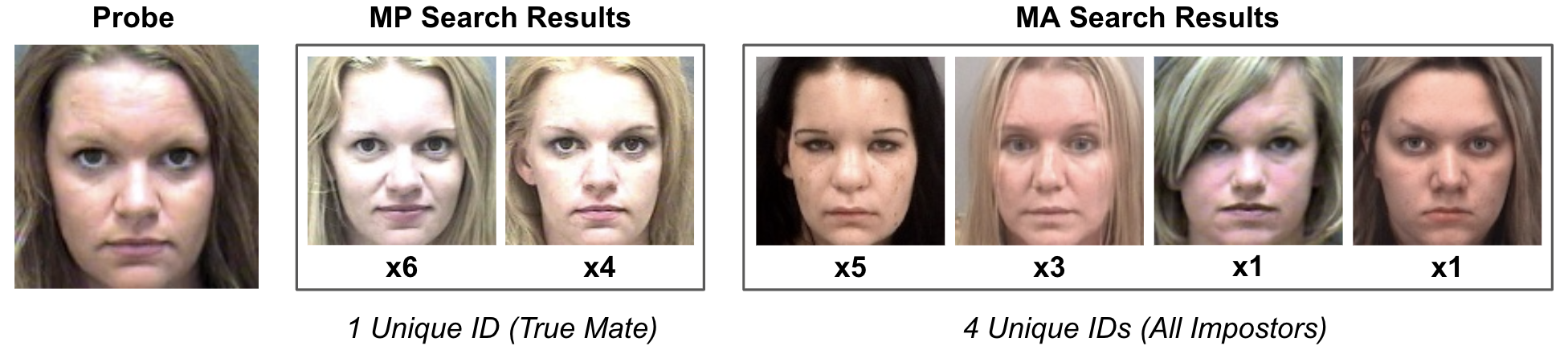}
\caption{Example rank consensus for a probe run in an MP search (two mates in gallery) versus MA search (no mates). ``Results'' show rank-1 images returned by the 10 matchers, where xN indicates the number of matchers that returned a given image.}
\label{fig:rankconsensus-intuition}
\end{figure*}

We therefore introduced 1-consistency, the first and only method based on rank-1 identity consensus across multiple independently trained matchers, rather than the scores of a single matcher~\cite{pangelinan2026rank1}. The underlying intuition is: a mate-present probe should produce a stable rank-1 prediction, with its true gallery mate consistently rising to the top across matchers. A mate-absent probe should be less stable, with each matcher's unique embedding space surfacing a different gallery impostor. (See example in Fig. \ref{fig:rankconsensus-intuition}.) The $k$-consistency rule, then, labels a probe MP if $M$ matchers return $\leq k$ unique identities; 1-consistency is the $k=1$ case. When compared against two fixed-threshold methods across configurations varying image quality and gallery composition, 1-consistency consistently had the highest MP recall and overall accuracy and the lowest demographic disparities.

This work extends that analysis. The remainder of the paper is organized as follows. Sec.~\ref{sec:related} reviews prior MP/MA classification methods. Sec.~\ref{sec:experiment-design} describes the expanded set of matching configurations. Sec.~\ref{sec:characterization} characterizes $k$-consistency performance as consensus variables $M$ and $k$ vary (previously fixed at $M=10$ and $k=1$). Sec.~\ref{sec:threshold-methods} establishes the two threshold-based methods, against which 1-consistency is compared in Sec.~\ref{sec:comparison}.

\subsection{Key Contributions}
\begin{itemize}

\item \textbf{Evaluation of Varied Matching Configurations.} We conduct the performance comparison and consensus characterization experiments across 36 (gallery, probe quality) scenarios. Two gallery families isolate the effects of image depth (1--4 images per identity) and identity depth (2,500--12,500 enrolled identities). Four probe quality settings—original, moderate blur ($\sigma$=4) or resolution reduction (18×18), high resolution reduction (14$\times$14)—isolate the effects of degradation type versus severity.

\item \textbf{Characterization of Rank Consensus Variables.} We characterize the cross-matcher consensus signal across both the number of matchers $M$ (from 2--10) and the consensus value $k$ (from 1--10). For degraded probes, increasing $M$ increases MA recall but decreases MP recall, while increasing $k$ has the opposite effect. We also observe that for 1-consistent probes ($k=1$), the agreed-upon gallery identity is nearly always the correct mate: the rank-1 TPIR is $\sim100\%$ for original-quality or medium degradation, and only drops to 97\% for maximum degradation.  

\item \textbf{Comparison to Realistic and Best-Case Thresholds.} We compare 1-consistency to two threshold-based methods that optimize Equal Error Rate (EER). Fixed Score-Thresholding (FST), calibrated once in baseline conditions, is included to represent a realistic deployment. FST performance degrades drastically as conditions diverge from the calibration point. Oracle Score-Thresholding (OST), re-calibrated for each (gallery, probe quality) scenario, represents the best-case thresholding deployment---but is purely theoretical, as it would be impossible in practice. For degraded probes, 1-consistency offers the highest accuracy, beating FST by 22.1 pp on average and matching OST within a fraction of a point, with no tuning required. 

\end{itemize}

\section{Related Work}
\label{sec:related}

Open-set recognition (OSR) assumes incomplete knowledge of the world at
training time: unknown classes may appear during
testing~\cite{scheirer2013toward}. Scheirer et al. posed OSR as a
constrained risk-minimization problem and introduced score-calibration
methods grounded in extreme value theory
(EVT)~\cite{scheirer2011metarecognition, scheirer2013toward,
scheirer2014probability}. Bendale and Boult extended these ideas to
incremental open-world learning~\cite{bendale2015towards} and deep
networks~\cite{bendale2016towards}. General OSR literature,
developed largely for object recognition, is now extensive; see ~\cite{geng2021survey} for an overview.

Open-set \emph{face} identification, by contrast, has received little dedicated attention---even though
transduction-based~\cite{li2005openset} and
visitor-interface~\cite{ekenel2009openset} approaches predate the
general formalizations above---and so has no standard terminology or
protocol. NIST evaluations frame it as mated versus non-mated search
and report two error types: false positives, when a probe with no
gallery mate is matched to an enrollee, and misses, when an enrolled
probe's mate is not returned; these are measured by the False Positive
and False Negative Identification Rates (FPIR and
FNIR)~\cite{FRVT_Identification, frvt-demographic}. Other work refers to
known unknowns versus unknown unknowns~\cite{guenther2017toward}, and our own
recent work used in-gallery and
out-of-gallery~\cite{pangelinan2026rank1, bhatta2025ingallery}, which we
now term MP and MA, respectively. In the
absence of a canonical taxonomy, we group prior work by its general
methodological approach.

The most direct approach thresholds a similarity score: a probe whose
top score falls below a fixed value is labeled
MA~\cite{FRVT_Identification, frvt-demographic}. This is simple but
brittle, because the ideal threshold shifts with the matcher and operational/demographic variables and so must be re-derived
for each deployment. A large body of work therefore replaces the raw
score with a calibrated quantity. EVT models the tail of the score
distribution to estimate the probability that a top match comes from a
genuine rather than an impostor identity:
meta-recognition~\cite{scheirer2011metarecognition}, the 1-vs-set
machine~\cite{scheirer2013toward}, and the CAP and W-SVM
models~\cite{scheirer2014probability} all produce such calibrated
decision boundaries, and the Extreme Value
Machine~\cite{rudd2017extreme} and later operational
refinements~\cite{cruz2024operational} continue this line. G\"unther et
al. show that thresholding EVM probabilities outperforms thresholding
verification-like scores on an LFW-based open-set
protocol~\cite{guenther2017toward}. Even these methods, however, still
depend on a decision boundary placed on a score-derived quantity, and
so can be sensitive to the same operational drift in quality and
gallery composition; their calibration rests on distributional
assumptions that may not hold across deployment conditions.

Deep open-set methods instead build rejection into the network.
OpenMax replaces the softmax layer with an EVT-calibrated estimate of
the unknown-class probability~\cite{bendale2016towards}, and
maximal-entropy and objectosphere losses train the network to respond
with low confidence to unknowns~\cite{vareto2024openset}. These improve
robustness but still calibrate a score-derived decision, and they
require training with curated known-unknown examples. Such data is
difficult to assemble representatively by construction, since the
unknowns a system meets in deployment are precisely the ones it did not
see during training~\cite{guenther2017toward, bendale2016towards}.

A related line improves identification robustness by aggregating the
multiple enrolled images of an identity rather than by rejecting
unknowns. These range from simple averaging of embeddings to learned
fusion: heterogeneous feature fusion~\cite{bodla2017deep}, neural
aggregation~\cite{yang2017neural}, cluster-and-aggregate for large
probe sets~\cite{kim2022caface}, probabilistic
embeddings~\cite{shi2019probabilistic}, and sparse-expert
fusion~\cite{jawade2024proxyfusion}. They are typically evaluated on
unconstrained benchmarks such as IJB-A~\cite{klare2015ijba},
IJB-B~\cite{whitelam2017ijbb}, and IJB-S~\cite{kalka2018ijb}, and
reliably raise accuracy under pose and quality variation. But they
address the accuracy of the match, not the enrollment decision, and the
fused score is still subject to the limits of thresholding.

Our recent works leverage rank information.
In~\cite{bhatta2025ingallery}, we classify a probe as MP or
MA using the ranks of the other enrolled images of the rank-1
identity, training a classifier on these rank-feature vectors. This approach
avoids distributional assumptions and is agnostic to the MP/MA mix, but
it needs more than one enrolled image per identity and still trains a
dedicated classifier. Our other prior work introduced rank-1 identity
consistency across independently trained
matchers~\cite{pangelinan2026rank1}, which the present paper extends.

We know of no work other than our own~\cite{pangelinan2026rank1} that
uses matcher consensus to classify a 1:N matching result as MP or MA.
Additionally, while several of the works above evaluate on
unconstrained imagery,
only ours \cite{bhatta2025ingallery, pangelinan2026rank1} characterize performance as a controlled function of probe characteristics.
For in-the-wild datasets, quality factors co-occur and cannot be varied independently, so a change in
performance cannot be attributed to any single cause. Our evaluation
instead applies controlled degradations to a fixed set of images,
following the approach used by NIST \cite{NIST-FATE-Report}. This isolates the
effect of each degradation type and severity while holding gallery size and composition fixed.

\section{Matching Configurations}
\label{sec:experiment-design}

The matching configurations described in this section are used across characterization and comparison experiments. Each is defined by two variables: the gallery being searched against and the quality of the probe image.

Probe and gallery images are from MORPH v5 ~\cite{ricanek2006morph, morph_site}: mugshot-quality
images captured under controlled conditions (frontal pose, neutral
expression, consistent lighting) and annotated with ground-truth
demographic information. 

We curate two non-overlapping subsets of the Black Male (BLM) cohort, totaling 12{,}926 unique identities. The ``probe set'' contains 2{,}123 identities with 5+ images; each identity's most recent image is its probe and its gallery mate(s) are selected from among older images. The ``gallery set'' contains 10{,}803 identities with 4+ images; these populate the gallery as needed. 

\subsection{Gallery Families}
\label{sec:experiment-design-galleries}

We construct two gallery families that isolate axes of composition variation. The first axis is image depth $x$---number of enrolled images per identity. The second axis is identity depth $y$---number of identities enrolled in the gallery.

\begin{figure}[t]
\centering
\includegraphics[width=\columnwidth]{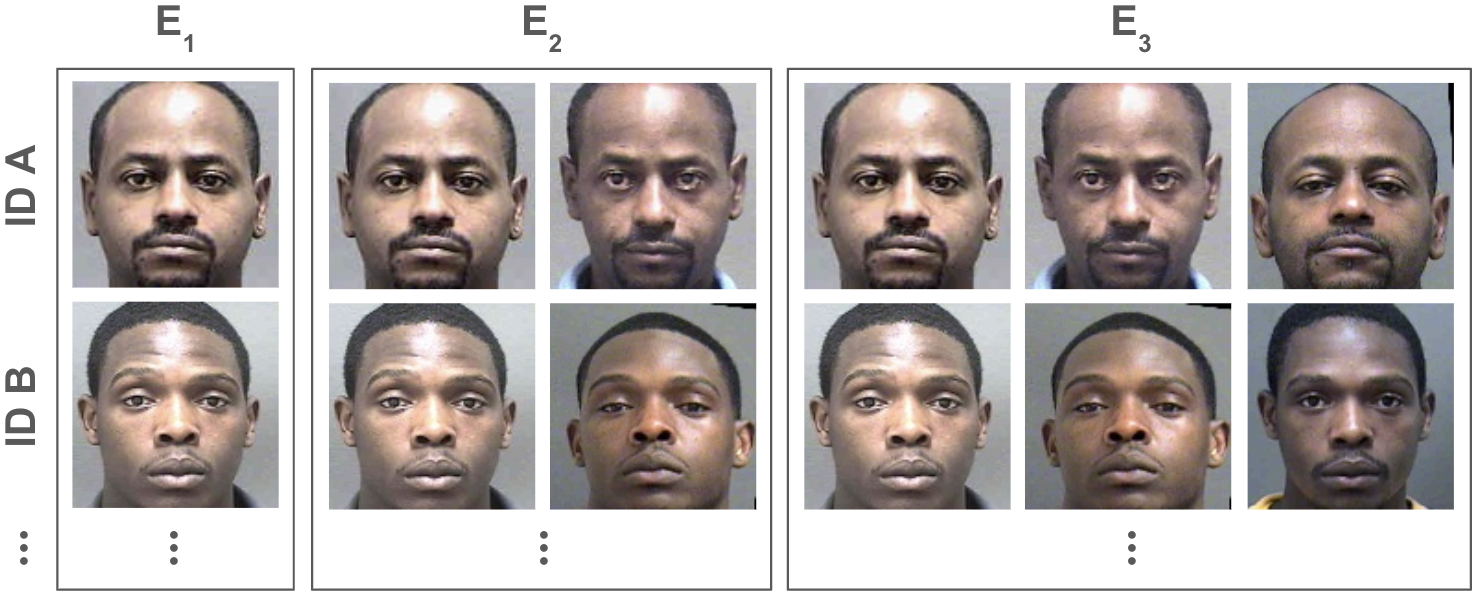}
\caption{Example image depth galleries $E_1-E_3$.}
\label{fig:num_images_experiment}
\end{figure}

In the image depth galleries (denoted $E_x$), we fix identity depth ($y=12{,}926$) and systematically increase $x$ from 1 to 4. Each gallery $E_x$ is a strict superset of the previous one (see Fig.~\ref{fig:num_images_experiment}), yielding the following total image counts: 12{,}926 ($E_1$), 25{,}852 ($E_2$), 38{,}778 ($E_3$), and 51{,}704 ($E_4$).

In the identity depth galleries (denoted $I_y$), we fix image depth ($x=2$, the average image depth used by NIST in~\cite{FRVT_Identification}) and systematically increase $y$ from 2{,}500 to 12{,}500 in steps of 2{,}500. The two images used for each identity correspond to those in the $E_2$ experiment. 

\begin{figure}[t]
\centering

\begin{subfigure}{0.22\columnwidth}
    \centering
    \includegraphics[width=\linewidth]{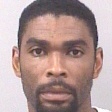}
    \caption{ORG}
\end{subfigure}
\hfill
\begin{subfigure}{0.22\columnwidth}
    \centering
    \includegraphics[width=\linewidth]{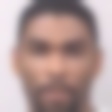}
    \caption{MS-B}
\end{subfigure}
\hfill
\begin{subfigure}{0.22\columnwidth}
    \centering
    \includegraphics[width=\linewidth]{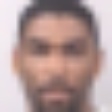}
    \caption{MS-R}
\end{subfigure}
\hfill
\begin{subfigure}{0.22\columnwidth}
    \centering
    \includegraphics[width=\linewidth]{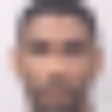}
    \caption{HS-R}
\end{subfigure}

\caption{Example image and quality-degraded variants.}

\label{fig:degradation_examples}

\end{figure}

\subsection{Probe Quality Levels}
\label{sec:experiment-design-degradations}

Starting with the original good-quality MORPH v5 images (abbreviated ORG), we apply two kinds of degradations to represent low-quality operational probes.  

Gaussian blur simulates degradation due to defocus. The
$\sigma$ parameter controls blur severity, with higher values
indicating more severe degradation. NIST quality testing
uses a range of $\sigma = 1$--$7$~\cite{NIST-FATE-Report}; we use the middle value $\sigma = 4$ as the 
moderate severity blur (MS-B) condition.

Downsampling--upsampling simulates native low-resolution capture. We use $18 \times 18$ for a moderate severity resolution degradation (MS-R), as prior work~\cite{pangelinan2026dual} showed its impact on recognition to be roughly equivalent to the selected $\sigma = 4$ blur. We use $14 \times 14$ for a high severity resolution degradation (HS-R). The corresponding interpupillary distances (pixels between the eyes) are approximately 6 px for MS-R and 5 px for HS-R.

Comparing MS-B and MS-R isolates the effect of degradation type; comparing MS-R and HS-R isolates the effect of severity. Fig.~\ref{fig:degradation_examples} shows the relative quality of the
original (a) and degraded (b--d) images.

\vspace{1em}
The nine galleries and four quality levels yield 36 unique (gallery, probe quality) scenarios. To keep results concise and readable, we adopt two presentation conventions. First, because gallery composition affects performance far less than probe quality (as we show throughout), many tables and figures report only the ``bookend'' galleries: $E_1$ and $E_4$ for the image depth family, $I_{2500}$ and $I_{12500}$ for the identity depth gallery. Second, because degradation \emph{type} matters far less than \emph{severity} (likewise shown throughout), we often omit MS-B and report ORG, MS-R, and HS-R to capture the severity range.

\section{Characterizing Rank Consensus}
\label{sec:characterization}

In this section, we isolate each consensus variable in turn: number of matchers $M$ and consensus value $k$---the maximum number of unique rank-1 identities for an MP probe. Our previous work fixed $M=10$ and $k=1$. Here, we first fix $k=1$ and measure how 1-consistency performance varies as $M$ increases from 2 to 10. Then, fixing $M=10$, we vary $k$ from 1 to 10 to examine its effect on classification.

\subsection{Varying the Number of Matchers $M$}   %
The matchers used here (and in \cite{pangelinan2026rank1}) are 10 independently trained instances of AdaFace \cite{kim2022adaface} from \cite{bhatta2025deep}; AdaFace is chosen because it is particularly suited to low-quality images. The instances share a backbone (ResNet100 \cite{resnet100}) and training set (WebFace4M \cite{zhu2021webface260m}) and differ only in their starting weights. Using multiple matchers raises a natural question: how does the number of matchers affect performance? Would $M<10$ yield the same results, and would $M>10$ improve them?

\begin{figure*}[tb]
\centering
\includegraphics[width=.85\textwidth]{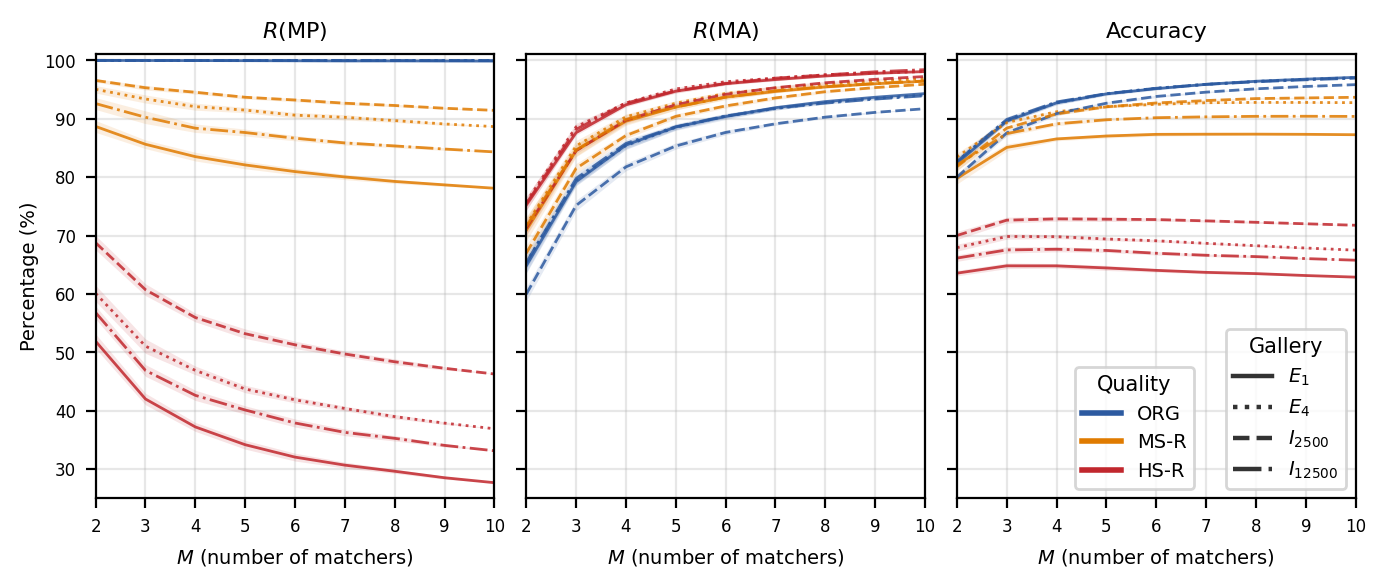}
\caption{1-consistency performance versus the number of matchers $M$, averaged over 20 random size-$M$  matcher subsets.}
\label{fig:n_matchers_curves}
\end{figure*}

We evaluate 1-consistency across size-$M$ subsets of the 10 matchers, for $M \in \{2, \dots, 10\}$. At each $M$ we draw 20 random subsets without replacement and report the mean and standard deviation of MP recall ($R(\text{MP})$), MA recall ($R(\text{MA})$), and accuracy (which is simply the average of the two). Fig.~\ref{fig:n_matchers_curves} shows overall results, with performance at three $M$ values highlighted in Tab. \ref{tab:n_matchers_net_change}.

As $M$ increases:

\begin{itemize}
    \item R(MP) remains consistently near 100\% for ORG probes, but decreases for degraded probes by 5--11 pp for MS-R and 22--24 pp for HS-R.
    \item R(MA) increases at a similar rate across quality levels (by roughly 29--32 pp for ORG, 25--29 pp for MS-R, 22--26 pp for HS-R), though its starting value at $M=2$ is higher for more degraded probes (60--65\% for ORG up to 71--76\% for HS-R).
    \item Accuracy continually increases for ORG probes (by 14--16 pp, still rising at $M=10$), plateaus for MS-R probes (around $M=8$), and for HS-R probes peaks early ($M=3-4$) before declining slightly (1--2 pp) through $M=10$.
\end{itemize}

Overall, curve shapes couple tightly with probe quality and are relatively consistent across galleries. However, the starting points for degraded probes (at $M=2$) reveal a consistent gallery ordering. R(MP) orders the galleries $I_{2500} > E_4 > I_{12500} > E_1$ at both degraded quality levels. R(MA) shows the inverse: $E_4, I_{12500},$ and $E_1$ cluster higher while $I_{2500}$ sits visibly lower. Accuracy reflects the balance of the two---it follows the R(MP) ordering at HS-R, but at MS-R the low R(MA) of $I_{2500}$ pulls it down to third. This inversion reflects gallery size: with the fewest enrolled identities, $I_{2500}$ offers the least opportunity for a lookalike to outrank a true mate (raising R(MP)) but also the fewest chances for matchers to disagree on an impostor (lowering R(MA)).

No single $M$ is optimal everywhere, so the choice is a trade-off. A deployment can tune $M$ for better R(MA), reducing wasted investigative effort, or R(MP), reducing missed leads---higher $M$ favors the former, lower $M$ the latter.

\begin{table}[t]
\centering
\caption{Performance at $M{=}2, 4, 10$.
\textbf{\textcolor{green!50!black}{Bold green}} indicates the best value
across the three $M$ per metric.}
\label{tab:n_matchers_net_change}
\resizebox{\columnwidth}{!}{%
\begin{tabular}{ll|ccc|ccc|ccc}
\toprule
 & & \multicolumn{3}{c|}{$R(\text{MP})$ @ $M{=}$} 
 & \multicolumn{3}{c|}{$R(\text{MA})$ @ $M{=}$} 
 & \multicolumn{3}{c}{Acc @ $M{=}$} \\
\cmidrule(lr){3-5}\cmidrule(lr){6-8}\cmidrule(lr){9-11}
\textbf{Gal} & \textbf{Probe} 
& \textbf{2} & \textbf{4} & \textbf{10} 
& \textbf{2} & \textbf{4} & \textbf{10} 
& \textbf{2} & \textbf{4} & \textbf{10} \\
\midrule

\multirow{3}{*}{\rotatebox[origin=c]{90}{$E_1$}}
& ORG 
& \textbf{\textcolor{green!50!black}{100}} 
& 99.9 
& 99.9 
& 64.8 
& 85.5 
& \textbf{\textcolor{green!50!black}{94.3}} 
& 82.4 
& 92.7 
& \textbf{\textcolor{green!50!black}{97.1}} \\

& MS-R 
& \textbf{\textcolor{green!50!black}{88.7}} 
& 83.5 
& 78.1 
& 70.9 
& 89.6 
& \textbf{\textcolor{green!50!black}{96.4}} 
& 79.8 
& 86.5 
& \textbf{\textcolor{green!50!black}{87.3}} \\

& HS-R 
& 51.9 
& 37.2 
& 27.7 
& 75.2 
& 92.4 
& \textbf{\textcolor{green!50!black}{98.1}} 
& 63.6 
& \textbf{\textcolor{green!50!black}{64.8}} 
& 62.9 \\

\midrule

\multirow{3}{*}{\rotatebox[origin=c]{90}{$E_4$}}
& ORG 
& \textbf{\textcolor{green!50!black}{100}} 
& \textbf{\textcolor{green!50!black}{100}} 
& \textbf{\textcolor{green!50!black}{100}} 
& 65.4 
& 85.4 
& \textbf{\textcolor{green!50!black}{94.1}} 
& 82.7 
& 92.7 
& \textbf{\textcolor{green!50!black}{97.0}} \\

& MS-R 
& \textbf{\textcolor{green!50!black}{95.0}} 
& 92.1 
& 88.6 
& 71.8 
& 90.3 
& \textbf{\textcolor{green!50!black}{96.8}} 
& 83.4 
& 91.2 
& \textbf{\textcolor{green!50!black}{92.7}} \\

& HS-R 
& 60.1 
& 46.9 
& 36.9 
& 75.7 
& 92.7 
& \textbf{\textcolor{green!50!black}{98.1}} 
& 67.9 
& \textbf{\textcolor{green!50!black}{69.8}} 
& 67.5 \\

\midrule

\multirow{3}{*}{\rotatebox[origin=c]{90}{$I_{2500}$}}
& ORG 
& \textbf{\textcolor{green!50!black}{100}} 
& \textbf{\textcolor{green!50!black}{100}} 
& \textbf{\textcolor{green!50!black}{100}} 
& 60.0 
& 81.8 
& \textbf{\textcolor{green!50!black}{91.7}} 
& 80.0 
& 90.9 
& \textbf{\textcolor{green!50!black}{95.8}} \\

& MS-R 
& \textbf{\textcolor{green!50!black}{96.6}} 
& 94.5 
& 91.4 
& 67.0 
& 87.1 
& \textbf{\textcolor{green!50!black}{95.9}} 
& 81.8 
& 90.8 
& \textbf{\textcolor{green!50!black}{93.7}} \\

& HS-R 
& 68.8 
& 56.0 
& 46.3 
& 71.3 
& 89.7 
& \textbf{\textcolor{green!50!black}{97.2}} 
& 70.0 
& \textbf{\textcolor{green!50!black}{72.9}} 
& 71.8 \\

\midrule

\multirow{3}{*}{\rotatebox[origin=c]{90}{$I_{12500}$}}
& ORG 
& \textbf{\textcolor{green!50!black}{100}} 
& \textbf{\textcolor{green!50!black}{100}} 
& \textbf{\textcolor{green!50!black}{100}} 
& 65.3 
& 85.8 
& \textbf{\textcolor{green!50!black}{93.9}} 
& 82.7 
& 92.9 
& \textbf{\textcolor{green!50!black}{96.9}} \\

& MS-R 
& \textbf{\textcolor{green!50!black}{92.6}} 
& 88.4 
& 84.3 
& 71.6 
& 89.9 
& \textbf{\textcolor{green!50!black}{96.5}} 
& 82.1 
& 89.1 
& \textbf{\textcolor{green!50!black}{90.4}} \\

& HS-R 
& 56.8 
& 42.6 
& 33.2 
& 75.5 
& 92.7 
& \textbf{\textcolor{green!50!black}{98.4}} 
& 66.1 
& \textbf{\textcolor{green!50!black}{67.7}} 
& 65.8 \\

\bottomrule
\end{tabular}
}
\end{table}

\subsection{Varying the Consensus Value $k$}       %

\begin{table}[t]
\centering
\caption{Mean count of unique rank-1 identities per probe.}
\label{tab:k_distribution_bookend}
\begin{tabular}{l|ccc|ccc}
\toprule
 & \multicolumn{3}{c|}{\textbf{Mean, MP Probes}} & \multicolumn{3}{c}{\textbf{Mean, MA Probes}} \\
\cmidrule(lr){2-4}\cmidrule(lr){5-7}
\textbf{Gal} & ORG & MS-R & HS-R & ORG & MS-R & HS-R \\
\midrule
$E_1$      & 1.00 & 1.53 & 3.57 & 4.33 & 4.94 & 5.35 \\
$E_4$      & 1.00 & 1.23 & 3.11 & 4.37 & 4.98 & 5.44 \\
$I_{2500}$ & 1.00 & 1.15 & 2.52 & 3.90 & 4.49 & 4.90 \\
$I_{12500}$& 1.00 & 1.34 & 3.33 & 4.35 & 4.96 & 5.40 \\
\bottomrule
\end{tabular}
\end{table}

Recall we have used $k=1$; one might ask what happens at higher values. Tab.~\ref{tab:k_distribution_bookend} shows the mean number of unique rank-1 identities for MP and MA probes across configurations. For both probe types the mean rises as quality degrades, but the two move differently. MA counts rise gradually and modestly (about +0.6 then +0.4 across the two degradation steps). MP counts barely move from ORG to MS-R (about +0.3) but jump sharply at HS-R (about +1.8), as severe degradation finally breaks the cross-matcher agreement that holds for milder degradation. The MP and MA means stay well separated at each quality level, though the gap narrows mainly at HS-R.

Within any single (gallery, probe quality) configuration, the MP and MA means are cleanly separated---a $k$ value can be chosen that splits them. But an operational deployment must handle images of varying quality, and no single $k$ separates them all: a very low-quality MP probe can produce as many unique identities as a good-quality MA probe, so their counts overlap. 

\begin{figure*}[h]
\centering
\includegraphics[width=.85\textwidth]{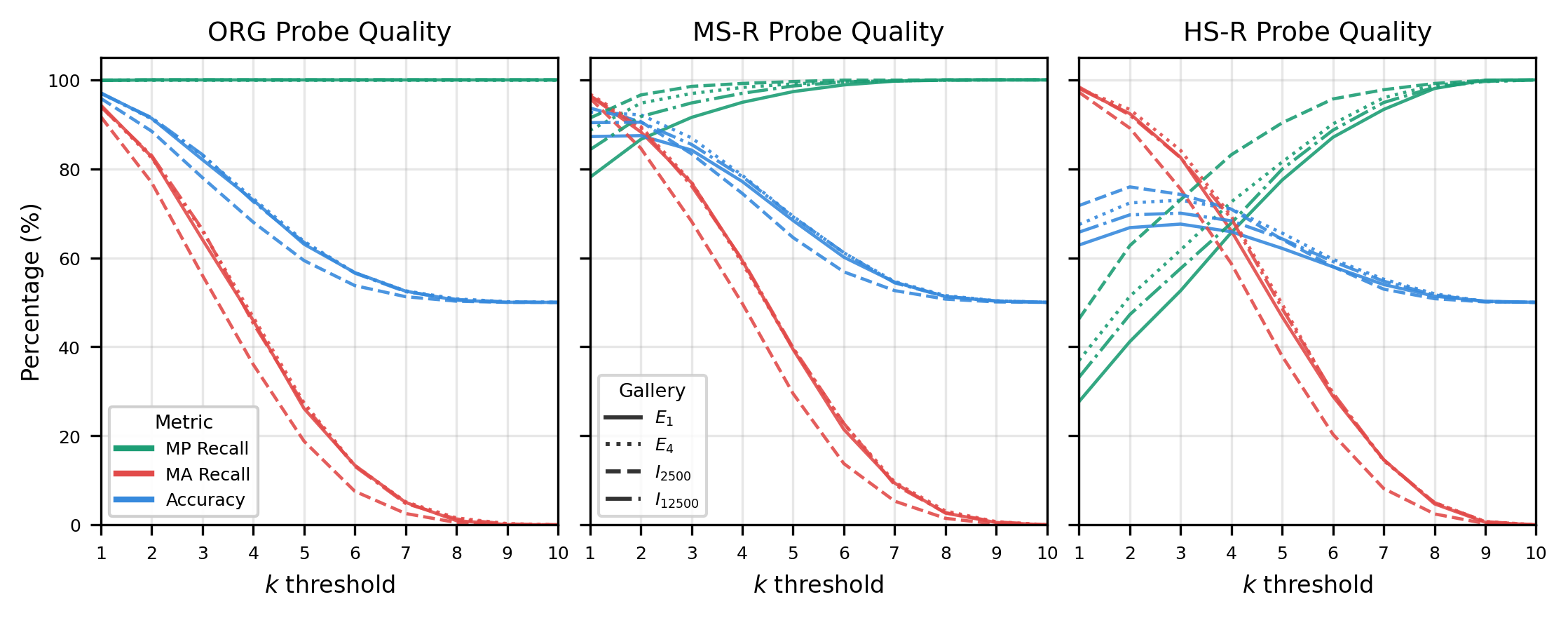}
\caption{Rank-1 identity consensus performance at different $k$ values.}
\label{fig:ksweep_bookends}
\end{figure*}

To see how this plays out across the full range of $k$ values, we sweep $k$ from 1 to 10 (Fig.~\ref{fig:ksweep_bookends}). As probe quality decreases (left to right), the $k$ that maximizes accuracy and the $k$ that equalizes MP and MA recall (the Equal Error Rate, EER) diverge. For ORG probes the two coincide at $k=1$; for HS-R probes the EER-optimal $k$ is consistently one step above the accuracy-optimal $k$. As with $M$, $k$ trades the two recalls against each other: a higher $k$ admits more probes as MP, favoring R(MP), while a lower $k$ favors R(MA).

\section{Threshold Methods}
\label{sec:threshold-methods}

We now describe the score-thresholding methods we will evaluate 1-consistency against. We are not aware of a standardized operational protocol for MP/MA classification; NIST's evaluations~\cite{FRVT_Identification}, for example, characterize 1:N error rates (FPIR/FNIR) as a function of operating threshold rather than prescribing a fixed decision rule. We therefore evaluate against two score-thresholding regimes that bracket what a deployed system could achieve, one realistic and one best-case but theoretical.

Fixed Score-Thresholding (FST) sets a single threshold once and applies it unchanged to every scenario. This reflects the constraint any deployed score-thresholding system operates under: the threshold must be fixed before any probe is seen, as per-probe recalibration is not feasible operationally. It is also the regime of our prior work~\cite{pangelinan2026rank1}.

Oracle Score-Thresholding (OST) is re-tuned to the optimal threshold for each (gallery, probe quality) combination. It is not deployable---it presumes per-scenario knowledge of the MP and MA score distributions, the very labels the system exists to predict---but it bounds the best performance any score threshold could attain, serving as a ceiling against which both FST and 1C can be compared.

The same tension between accuracy-maximizing and error-balancing operating points observed for $k$ arises in setting the actual score threshold for either method. Equal Error Rate (EER) places the threshold where the MP and MA error rates are equal; Maximum Accuracy (MaxAcc) places it wherever overall accuracy is highest. We resolve the choice with respect to OST, where the threshold is optimal for each scenario by construction, so the comparison between criteria is not confounded by calibration error. 

Tab.~\ref{tab:eer_vs_maxacc_bookend} reports OST performance for EER and MaxAcc criteria. For ORG probes the two are effectively identical: the accuracy-maximizing threshold already balances the error rates, with an accuracy difference below 0.03 pp. The criteria diverge only under degradation. MaxAcc's accuracy advantage is small---at most 2.8 pp under HS-R, and roughly 0.9 pp averaged across all 36 scenarios. What it gives up for that advantage is not: to recover its last fraction of accuracy, MaxAcc drives the threshold until MP recall collapses---the R(MP)/R(MA) gap under HS-R runs 25 to 43 pp, with R(MP) as low as 46\% while R(MA) sits in the mid-to-high 80s---whereas EER holds the two recalls within 1.3 pp. The mechanism is direct: under severe degradation the MP score distribution shifts down into the MA range, so the only way to raise accuracy further is to reject nearly everything near the overlap, most of which is MP. 

Because the accuracy gained is marginal and the recall imbalance severe, we adopt EER for both OST and FST throughout. This also keeps FST consistent with the fixed-EER threshold of our prior work~\cite{pangelinan2026rank1}. FST is calibrated in the baseline ($E_2$, ORG) scenario; the single threshold is 0.495. The tuned thresholds for OST are given in Tab. \ref{tab:ost_thresholds}.

\begin{table}[t]
\centering
\caption{EER vs. MaxAcc calibration for OST.}
\label{tab:eer_vs_maxacc_bookend}
\resizebox{\columnwidth}{!}{%
\begin{tabular}{ll|ccc|ccc}
\toprule
& & \multicolumn{3}{c|}{EER} & \multicolumn{3}{c}{MaxAcc} \\
\cmidrule(lr){3-5}\cmidrule(lr){6-8}
\textbf{Gal} & \textbf{Probe} & \textbf{$R(\text{MP})$} & \textbf{$R(\text{MA})$} & \textbf{Acc} & \textbf{$R(\text{MP})$} & \textbf{$R(\text{MA})$} & \textbf{Acc} \\
\midrule
\multirow{2}{*}{$E_1$} & ORG & 99.6 & 99.6 & 99.6 & 99.6 & 99.6 & 99.6 \\
 & HS-R & 63.2 & 62.3 & 62.8 & 47.6 & 82.6 & 65.1 \\
\midrule
\multirow{2}{*}{$E_4$} & ORG & 99.8 & 99.8 & 99.8 & 99.8 & 99.8 & 99.8 \\
 & HS-R & 66.7 & 66.7 & 66.7 & 51.7 & 86.9 & 69.3 \\
\midrule
\multirow{2}{*}{$I_{2500}$} & ORG & 99.9 & 99.9 & 99.9 & 99.9 & 99.9 & 99.9 \\
 & HS-R & 71.8 & 72.1 & 71.9 & 60.3 & 85.9 & 73.1 \\
\midrule
\multirow{2}{*}{$I_{12500}$} & ORG & 99.8 & 99.8 & 99.8 & 99.7 & 99.9 & 99.8 \\
 & HS-R & 64.1 & 65.4 & 64.8 & 45.7 & 88.6 & 67.1 \\
\bottomrule
\end{tabular}%
}
\end{table}

\begin{table}[tb]
\centering
\caption{Per-scenario OST thresholds.}
\label{tab:ost_thresholds}
\begin{tabular}{l|cccc}
\toprule
\textbf{Gallery} & \textbf{Original} & \textbf{MS-Blur} & \textbf{MS-Res} & \textbf{HS-Res} \\
\midrule
$E_1$       & 0.455 & 0.352 & 0.357 & 0.310 \\
$E_2$       & 0.495 & 0.368 & 0.373 & 0.322 \\
$E_3$       & 0.515 & 0.376 & 0.383 & 0.328 \\
$E_4$       & 0.519 & 0.382 & 0.388 & 0.333 \\
\specialrule{1pt}{2pt}{2pt}
$I_{2500}$  & 0.472 & 0.347 & 0.351 & 0.296 \\
$I_{5000}$  & 0.495 & 0.358 & 0.362 & 0.308 \\
$I_{7500}$  & 0.495 & 0.362 & 0.367 & 0.314 \\
$I_{10000}$ & 0.495 & 0.365 & 0.370 & 0.318 \\
$I_{12500}$ & 0.495 & 0.368 & 0.373 & 0.322 \\
\specialrule{1pt}{2pt}{2pt}
\textbf{Mean}    & \textbf{0.488} & \textbf{0.366} & \textbf{0.371} & \textbf{0.316} \\
\textbf{Range}   & 0.064 & 0.035 & 0.037 & 0.037 \\
\bottomrule
\end{tabular}
\end{table}

\section{Comparing Classification Performance} \label{sec:comparison}

Finally, we compare the three methods across all scenarios. Tab.~\ref{tab:full_comparison_table} reports results with signed 1C--OST difference ($\Delta$ block); Fig.~\ref{fig:method_by_metric} shows the same comparison averaged within each gallery family. MS-Blur tracked MS-Res within 2--3 pp on every metric, so the figures follow the ORG $\rightarrow$ MS-R $\rightarrow$ HS-R severity ladder.

\subsection{1-Consistency versus the Oracle Score Threshold}
The accuracy comparison reverses with probe quality, and the reversal is the central result. For ORG probes OST leads 1C by a few points. The reason is in the component recalls: 1C's MP recall is saturated near 100\%, so accuracy is governed by MA recall, and 1C's MA recall at ORG is only about 92--94\%. The strict unanimity rule (i.e., $k=1$) occasionally rejects a clean MA probe whose ten matchers happen to disagree, and with nothing moving on the MP side, that lost MA recall accounts for the accuracy gap.

As soon as quality degrades the ordering inverts: 1C holds the best or tied-best accuracy in 25 of 27 degraded-probe scenarios. Under degradation, OST's MA recall falls steeply---to roughly 62--72\% at HS-R---as degraded MP scores intrude into the range OST must accept, dragging MA probes with them. 1C's MA recall, by contrast, stays near 97--98\% throughout, because the matchers are rarely unanimous on an MA probe's rank-1 identity regardless of where its scores fall. 

The 1C--OST MA margin scales with severity: as small as 1.2 pp under the mildest degradation and as large as 35.8 pp at the most severe. The ordering holds across every gallery in both families---at fixed quality the spread across galleries is a few points, while across quality within a fixed gallery it is tens of points.

In accuracy, the two methods are nearly indistinguishable under degradation: $\Delta$Acc stays within 1.3 pp of zero across all degraded scenarios, as 1C's MP loss and MA gain relative to OST very nearly cancel. The Accuracy panel of Fig.~\ref{fig:method_by_metric} shows this directly---1C and OST track together across the ladder. What separates them is not how often they err but which error they make: the $\Delta$ block shows $\Delta$R(\text{MP}) negative and $\Delta$R(MA) positive under degradation, of similar magnitude, with $\Delta$Acc near zero between them.

\subsection{Fixed Score-Thresholding}
\begin{figure}[t]
\centering
\includegraphics[width=\columnwidth]{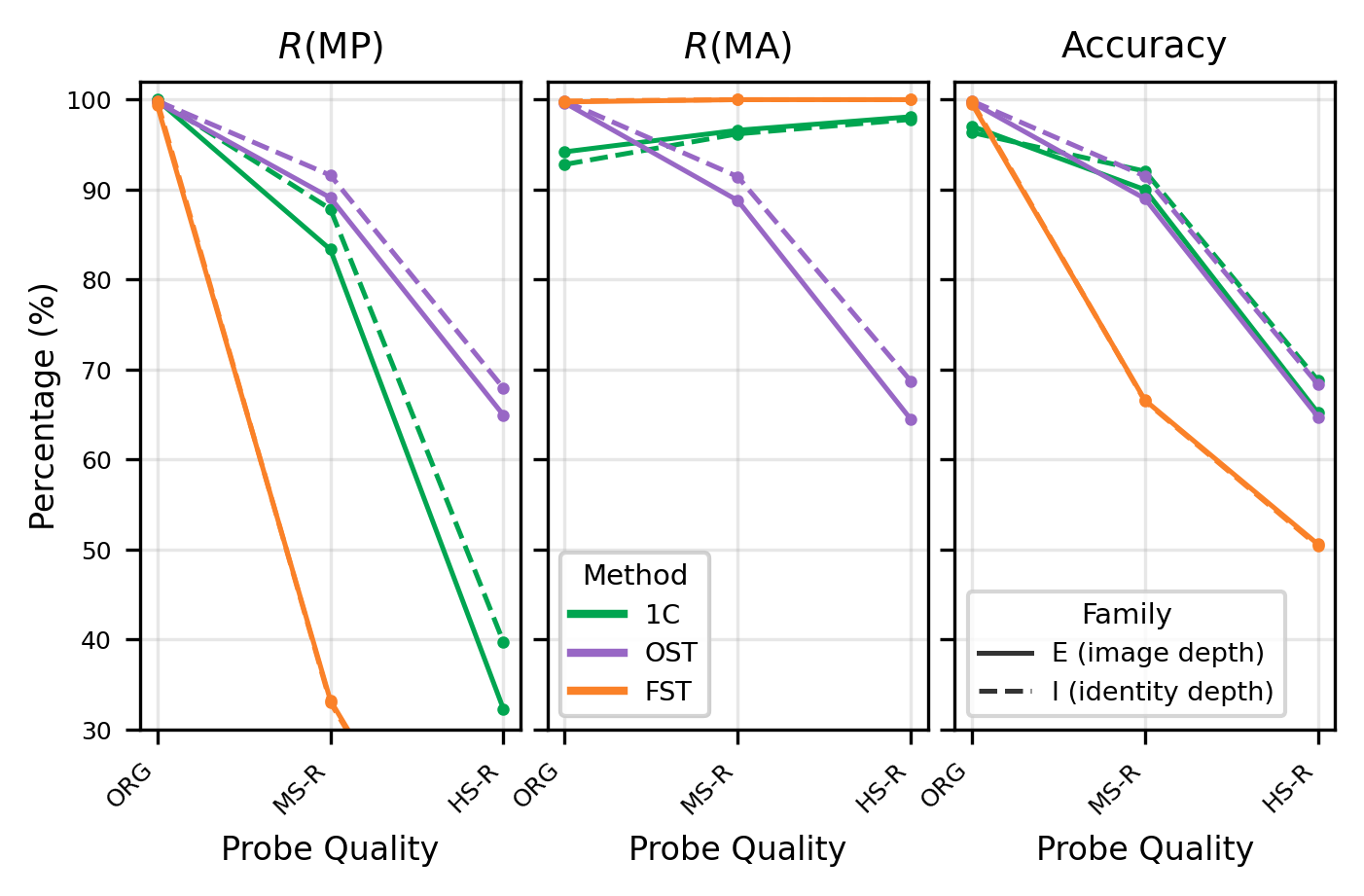}
\caption{Comparison of the three classification approaches.}
\label{fig:method_by_metric}
\end{figure}

FST applies its single threshold of $s=0.495$, calibrated in the ($E_2$, ORG) scenario, unchanged everywhere. For probes matching the calibration point it performs comparably---ORG accuracy within about a point of 1C and OST. As quality moves away from the calibration point the fixed threshold sits far above the degraded MP distribution, and the consequence is severe: at HS-R, FST's MP recall falls to between 0.6\% and 1.7\%---it rejects very nearly every genuine mate---and its accuracy drops by up to 49.4 pp relative to its own ORG performance. In Fig.~\ref{fig:method_by_metric}, the FST MP line falls off the bottom of the axis.

\subsection{Identity Correctness of MP Decisions}
\begin{figure}[t]
\centering
\includegraphics[width=\columnwidth]{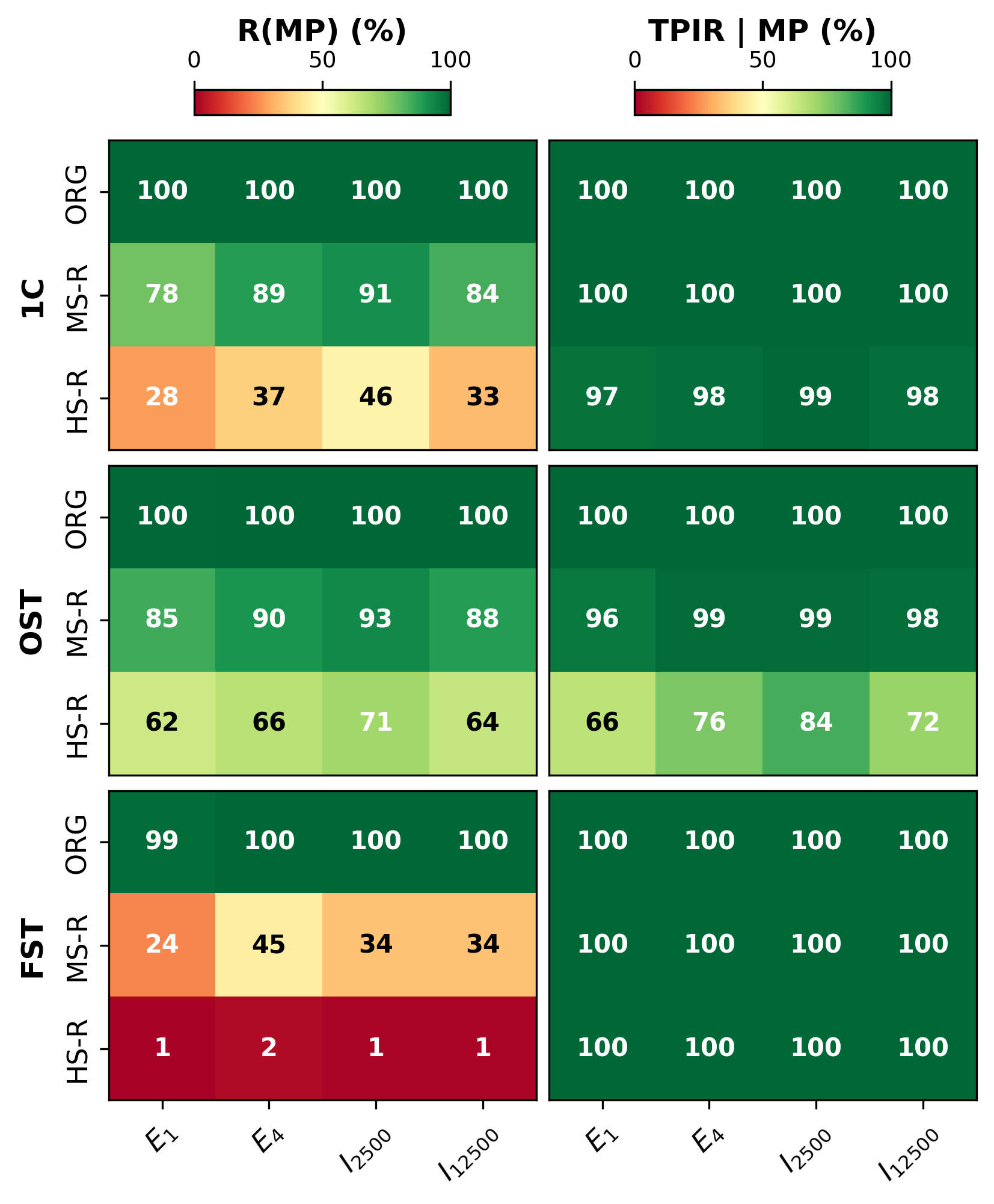}
\caption{R(MP) (left) and TPIR conditional on an MP label (right) for the three methods.}
\label{fig:tpir_heatmaps}
\end{figure}

The comparison so far asks whether a probe with a gallery mate is labeled MP. A separate question remains: when a method does label a probe MP, is the identity it returns the \textit{correct} mate? We measure this as the true positive identification rate (TPIR) conditional on an MP label---of the MP-labeled probes, the fraction whose returned identity is the true mate. 1-consistency is an ensemble decision (one label per probe across all M matchers), so its returned identity is the single consensus identity. The threshold methods are single-matcher decisions; consistent with the recalls reported above, we evaluate them per probe and per matcher, asking for each run whether a score above the threshold was attached to the correct mate. Fig.~\ref{fig:tpir_heatmaps} pairs R(MP) with this conditional TPIR for the three methods.

When 1-consistency labels a probe MP, the consensus identity is almost always the true mate: TPIR is 100\% for ORG probes and stays at 97--99\% even for HS-R. Unanimous agreement across ten matchers is rarely agreement on a wrong identity. OST is likewise near-perfect at ORG (100\%) and MS-R (96–99\%), but at HS-R its conditional TPIR falls to 66--84\%. Under severe degradation, a substantial share of the probes it accepts are matched to an \textit{impostor} rather than the true mate. FST's conditional TPIR is 100\% throughout, but this must be read against its R(MP): at HS-R it labels only 1--2\% of MP probes as MP, and the few probes that clear its high fixed threshold are precisely the unambiguous ones whose top match is correct. Its perfect identity correctness is therefore a consequence of accepting almost nothing, rather than a strength.

Read together, the two rows separate the methods by the character of their MP decisions. 1-consistency accepts a substantial fraction of degraded MP probes and is almost always right about who they are. OST accepts a comparable or larger fraction but, under severe degradation, is markedly less reliable about identity. FST accepts so few that the question of identity correctness is moot.

\section{Discussion}
\label{sec:discussion}

\noindent\textbf{When methods tie in accuracy, they differ in which error they make.} Any gallery-membership method trades MP recall against MA recall: pushing one up generally pulls the other down. Under degradation, 1-consistency holds the best or tied-best overall accuracy in 25 of 27 scenarios (the two exceptions trail the oracle by less than 0.3 pp). Relative to OST, it loses on MP recall but gains nearly the same on MA recall, so the two methods' accuracies never differ by more than 1.3 pp. 1-consistency errs toward missing a MP probe, while a score threshold errs increasingly toward accepting an MA one. Which error is more tolerable is deployment-specific, and is a choice the operator must make rather than one the method should impose.
\vspace{0.5em}

\noindent\textbf{No fixed threshold can generalize across quality.} A real deployment cannot recalibrate per probe. It would need to commit to threshold before any probe is seen, which is the constraint FST captures. When probe quality matches the calibration point, FST performs comparably to the other methods; when it does not, FST does not degrade gracefully but collapses, with MP recall falling to 0.6--1.7\% under HS-R and accuracy dropping up to 49.4 pp. The oracle explains why this is unavoidable rather than incidental. OST is the best any threshold could do given the ``optimal value'' for every scenario. Yet that threshold value ranges from 0.296--0.519 across our scenarios, varying almost entirely with probe quality (ORG thresholds occupy 0.46--0.52, HS-R 0.30--0.33) and only slightly across galleries. The threshold a deployment would need is thus a function of the operational conditions it encounters, which are not known in advance---they are \textit{unknown unknowns}.

\vspace{0.5em}

\noindent\textbf{When 1-consistency accepts a probe, it is more reliable about identity than the oracle score threshold.} Accuracy and recall measure whether the membership decision is right, but not whether the returned identity is. Conditional on labeling a probe mate-present, 1-consistency surfaces the true mate in 97--100\% of cases across all qualities, because ten matchers rarely agree unanimously on a wrong identity. The oracle, despite its per-scenario tuning, falls to 66--84\% under HS-R: a substantial share of the probes it accepts are high-scoring impostors rather than true mates. This is the one axis on which 1-consistency does not merely match the oracle but exceeds it---the oracle is a ceiling on accuracy, not on identity reliability. (FST's perfect conditional identity rate is not a counterexample: it accepts so few degraded probes, 1--2\% at HS-R, that the measure is moot.)
\vspace{0.5em}

\noindent\textbf{Probe quality outweighs the gallery variation we test.} Across every analysis, the dominant axis of variation is probe quality. For fixed quality, metrics vary by a few points across the nine galleries spanning both structural axes. For fixed gallery, the same metrics vary by tens of points across the four quality levels. This dominance holds over the ranges we vary, but should not be read as gallery structure being unimportant in general, because those ranges are narrower than operational reality on every axis.
Our largest gallery ($E_4$ with $\sim$51k images of $\sim$13k identities) is roughly three orders of magnitude smaller than that used by NIST in ~\cite{FRVT_Identification} ($\sim$26M images of $\sim$12M identities), and up to six orders of magnitude smaller than some commercial galleries---Clearview AI, for instance, reports over 70~\emph{billion} images in its law enforcement database~\cite{clearview_ai_2_0} (though not the number of identities this corresponds to). We also enroll every identity at a uniform image depth, whereas operational galleries enroll unevenly---some identities with many images, others with few---and this imbalance itself affects recognition~\cite{pangelinan2024analyzing}. Both axes we vary are thus academic samples of operational conditions: we can neither experiment at deployed gallery scales nor reproduce the mixed, co-occurring degradations of real probes. Our controlled single-degradation sweep isolates causes precisely because it does not attempt to mimic that uncontrolled mixture. Whether quality continues to dominate once these axes reach operational scale---where a larger impostor population gives MA probes more chances to draw spurious agreement on a single look-alike---remains open.
\vspace{0.5em}

\subsection{Limitations}
\begin{itemize}
    \item These results use synthetic degradations on good-quality imagery. Real surveillance probes additionally exhibit pose, occlusion, expression, and illumination variation, typically in combination.
    \item This study evaluates a single demographic cohort (BLM) and therefore does not test the cross-demographic fairness behavior we previously reported \cite{pangelinan2026rank1}.
    \item  Our ten matcher instances are homogeneous by construction: they share a backbone, training set, and loss, differing only in initialization. Consensus among more diverse matchers—varying the architecture, the training data (e.g., independently sampled subsets), or the loss—may behave differently, and a pool could also be assembled by selecting the most diverse matchers from a larger set. Whether such diversity strengthens or weakens the consensus signal is untested.
\end{itemize}

\section{Conclusion}
This paper extends the 1-consistency method for MP/MA classification in operational 1:N face identification~\cite{pangelinan2026rank1}. By separating gallery structure into image depth and identity depth, we show that probe quality, not gallery composition, drives classification performance across the configurations tested. Against both a fixed threshold set once (FST) and an oracle threshold tuned per scenario (OST), 1-consistency holds the best or near-best accuracy in nearly all degraded scenarios, with no calibration required. And when it accepts a probe, it identifies the correct mate more reliably than even the oracle (97--100\% versus 66--84\% under severe degradation). The advantage over a threshold is not higher accuracy so much as the removal of a commitment a deployment cannot honor: a single threshold cannot track the probe-quality regime it will encounter, while rank-1 consensus requires none over a wider range of probe quality.

This last point extends to operational reality. We are aware of no standard for the MP/MA decision---in the literature or in deployed practice---and operational systems may return rank-1 candidates with no explicit enrollment check at all. That absence is itself the risk this work addresses: an unscreened MA search yields a confident false match, with no signal that the returned identity is spurious. A method needing no threshold and no per-condition tuning lowers the barrier to making this decision at all, which may matter as much as which method is used.

Several directions remain open. The limitations above each point to a next step---evaluation on real surveillance probes rather than synthetic degradations, a consensus pool diversified in architecture, training data, or loss (or selected for diversity from a larger set) rather than instances of a single model, and larger-scale galleries on both the identity and image-depth axes. Beyond these, one direction is internal to the method: 1-consistency exposes two consensus controls, the number of matchers $M$ and the consensus value $k$, each trading MP against MA recall. We characterized each in isolation, but how they interact---and whether tuning them jointly widens the operating range a deployment can reach---is untested. Of the gallery-scale questions, the sharpest is whether probe quality continues to dominate as the impostor population grows: a larger gallery gives MA probes more opportunities to draw coincidental cross-matcher agreement on a single look-alike, the one regime in which 1-consistency's MA advantage could erode.

\begin{table*}[h]
\centering
\caption{Comparison of the three classification approaches. Per metric trio, \textbf{\textcolor{best}{bold green}} marks the best value and \textbf{\textcolor{worst}{bold red}} the worst; ties highlighted identically. The $\Delta$ block reports 1C minus OST for each metric (signed; positive favors 1C).}
\label{tab:full_comparison_table}
\resizebox{.9\textwidth}{!}{%
\begin{tabular}{ll|ccc|ccc|ccc|ccc}
\toprule
& & \multicolumn{3}{c|}{$R(\text{MP})$} & \multicolumn{3}{c|}{$R(\text{MA})$} & \multicolumn{3}{c|}{Acc} & \multicolumn{3}{c}{$\Delta$ (1C$-$OST)} \\
\cmidrule(lr){3-5}\cmidrule(lr){6-8}\cmidrule(lr){9-11}\cmidrule(lr){12-14}
\textbf{Gal} & \textbf{Probe} & \textbf{1C} & \textbf{OST} & \textbf{FST} & \textbf{1C} & \textbf{OST} & \textbf{FST} & \textbf{1C} & \textbf{OST} & \textbf{FST} & \textbf{$R(\text{MP})$} & \textbf{$R(\text{MA})$} & \textbf{Acc} \\
\midrule
\multirow{4}{*}{$E_1$} & ORG & \textbf{\textcolor{best}{99.9}} & 99.6 & \textbf{\textcolor{worst}{98.9}} & \textbf{\textcolor{worst}{94.3}} & 99.5 & \textbf{\textcolor{best}{99.8}} & \textbf{\textcolor{worst}{97.1}} & \textbf{\textcolor{best}{99.6}} & 99.4 & +0.3 & -5.2 & -2.5 \\
 & MS-B & 79.2 & \textbf{\textcolor{best}{87.5}} & \textbf{\textcolor{worst}{22.6}} & 97.6 & \textbf{\textcolor{worst}{87.8}} & \textbf{\textcolor{best}{100}} & \textbf{\textcolor{best}{88.4}} & 87.7 & \textbf{\textcolor{worst}{61.3}} & -8.2 & +9.8 & +0.8 \\
 & MS-R & 78.1 & \textbf{\textcolor{best}{86.6}} & \textbf{\textcolor{worst}{22.8}} & 96.4 & \textbf{\textcolor{worst}{86.3}} & \textbf{\textcolor{best}{100}} & \textbf{\textcolor{best}{87.3}} & 86.5 & \textbf{\textcolor{worst}{61.4}} & -8.5 & +10.1 & +0.8 \\
 & HS-R & 27.7 & \textbf{\textcolor{best}{63.2}} & \textbf{\textcolor{worst}{0.6}} & 98.1 & \textbf{\textcolor{worst}{62.3}} & \textbf{\textcolor{best}{100}} & \textbf{\textcolor{best}{62.9}} & 62.8 & \textbf{\textcolor{worst}{50.3}} & -35.5 & +35.8 & +0.1 \\
\cmidrule(lr){1-14}
\multirow{4}{*}{$E_2$} & ORG & \textbf{\textcolor{best}{100}} & \textbf{\textcolor{worst}{99.8}} & \textbf{\textcolor{worst}{99.8}} & \textbf{\textcolor{worst}{93.9}} & \textbf{\textcolor{best}{99.8}} & \textbf{\textcolor{best}{99.8}} & \textbf{\textcolor{worst}{96.9}} & \textbf{\textcolor{best}{99.8}} & \textbf{\textcolor{best}{99.8}} & +0.2 & -5.8 & -2.8 \\
 & MS-B & 86.3 & \textbf{\textcolor{best}{91.1}} & \textbf{\textcolor{worst}{33.0}} & 97.3 & \textbf{\textcolor{worst}{91.0}} & \textbf{\textcolor{best}{100}} & \textbf{\textcolor{best}{91.8}} & 91.1 & \textbf{\textcolor{worst}{66.5}} & -4.9 & +6.4 & +0.8 \\
 & MS-R & 84.2 & \textbf{\textcolor{best}{89.5}} & \textbf{\textcolor{worst}{33.0}} & 96.7 & \textbf{\textcolor{worst}{89.2}} & \textbf{\textcolor{best}{100}} & \textbf{\textcolor{best}{90.4}} & 89.4 & \textbf{\textcolor{worst}{66.5}} & -5.3 & +7.5 & +1.1 \\
 & HS-R & 33.0 & \textbf{\textcolor{best}{64.4}} & \textbf{\textcolor{worst}{0.9}} & 98.4 & \textbf{\textcolor{worst}{65.1}} & \textbf{\textcolor{best}{100}} & \textbf{\textcolor{best}{65.7}} & 64.8 & \textbf{\textcolor{worst}{50.4}} & -31.4 & +33.3 & +0.9 \\
\cmidrule(lr){1-14}
\multirow{4}{*}{$E_3$} & ORG & \textbf{\textcolor{best}{100}} & \textbf{\textcolor{worst}{99.8}} & \textbf{\textcolor{worst}{99.8}} & \textbf{\textcolor{worst}{94.2}} & \textbf{\textcolor{best}{99.8}} & 99.7 & \textbf{\textcolor{worst}{97.1}} & \textbf{\textcolor{best}{99.8}} & 99.7 & +0.2 & -5.6 & -2.7 \\
 & MS-B & 88.8 & \textbf{\textcolor{best}{92.1}} & \textbf{\textcolor{worst}{40.6}} & 97.5 & \textbf{\textcolor{worst}{92.0}} & \textbf{\textcolor{best}{100}} & \textbf{\textcolor{best}{93.1}} & 92.1 & \textbf{\textcolor{worst}{70.3}} & -3.3 & +5.4 & +1.1 \\
 & MS-R & 86.9 & \textbf{\textcolor{best}{90.5}} & \textbf{\textcolor{worst}{39.8}} & 96.7 & \textbf{\textcolor{worst}{90.9}} & \textbf{\textcolor{best}{100}} & \textbf{\textcolor{best}{91.8}} & 90.7 & \textbf{\textcolor{worst}{69.9}} & -3.6 & +5.8 & +1.1 \\
 & HS-R & 35.6 & \textbf{\textcolor{best}{66.0}} & \textbf{\textcolor{worst}{1.5}} & 98.3 & \textbf{\textcolor{worst}{65.6}} & \textbf{\textcolor{best}{100}} & \textbf{\textcolor{best}{66.9}} & 65.8 & \textbf{\textcolor{worst}{50.7}} & -30.4 & +32.7 & +1.1 \\
\cmidrule(lr){1-14}
\multirow{4}{*}{$E_4$} & ORG & \textbf{\textcolor{best}{100}} & \textbf{\textcolor{worst}{99.8}} & \textbf{\textcolor{worst}{99.8}} & \textbf{\textcolor{worst}{94.1}} & \textbf{\textcolor{best}{99.8}} & 99.7 & \textbf{\textcolor{worst}{97.0}} & \textbf{\textcolor{best}{99.8}} & 99.7 & +0.2 & -5.7 & -2.8 \\
 & MS-B & 90.3 & \textbf{\textcolor{best}{93.0}} & \textbf{\textcolor{worst}{45.3}} & 97.1 & \textbf{\textcolor{worst}{93.2}} & \textbf{\textcolor{best}{100}} & \textbf{\textcolor{best}{93.7}} & 93.1 & \textbf{\textcolor{worst}{72.6}} & -2.6 & +3.9 & +0.6 \\
 & MS-R & 88.6 & \textbf{\textcolor{best}{91.6}} & \textbf{\textcolor{worst}{43.7}} & 96.8 & \textbf{\textcolor{worst}{91.3}} & \textbf{\textcolor{best}{100}} & \textbf{\textcolor{best}{92.7}} & 91.5 & \textbf{\textcolor{worst}{71.8}} & -2.9 & +5.5 & +1.3 \\
 & HS-R & 36.9 & \textbf{\textcolor{best}{66.7}} & \textbf{\textcolor{worst}{1.7}} & 98.1 & \textbf{\textcolor{worst}{66.7}} & \textbf{\textcolor{best}{100}} & \textbf{\textcolor{best}{67.5}} & 66.7 & \textbf{\textcolor{worst}{50.9}} & -29.7 & +31.4 & +0.8 \\
\specialrule{1.5pt}{3pt}{3pt}
\multirow{4}{*}{$I_{2500}$} & ORG & \textbf{\textcolor{best}{100}} & 99.9 & \textbf{\textcolor{worst}{99.8}} & \textbf{\textcolor{worst}{91.7}} & \textbf{\textcolor{best}{99.9}} & \textbf{\textcolor{best}{99.9}} & \textbf{\textcolor{worst}{95.8}} & \textbf{\textcolor{best}{99.9}} & 99.8 & +0.1 & -8.1 & -4.0 \\
 & MS-B & 93.2 & \textbf{\textcolor{best}{94.6}} & \textbf{\textcolor{worst}{33.0}} & 95.9 & \textbf{\textcolor{worst}{94.7}} & \textbf{\textcolor{best}{100}} & 94.5 & \textbf{\textcolor{best}{94.7}} & \textbf{\textcolor{worst}{66.5}} & -1.4 & +1.2 & -0.1 \\
 & MS-R & 91.4 & \textbf{\textcolor{best}{93.7}} & \textbf{\textcolor{worst}{33.0}} & 95.9 & \textbf{\textcolor{worst}{93.5}} & \textbf{\textcolor{best}{100}} & \textbf{\textcolor{best}{93.7}} & 93.6 & \textbf{\textcolor{worst}{66.5}} & -2.3 & +2.4 & +0.1 \\
 & HS-R & 46.3 & \textbf{\textcolor{best}{71.8}} & \textbf{\textcolor{worst}{0.9}} & 97.2 & \textbf{\textcolor{worst}{72.1}} & \textbf{\textcolor{best}{100}} & 71.8 & \textbf{\textcolor{best}{71.9}} & \textbf{\textcolor{worst}{50.4}} & -25.5 & +25.2 & -0.2 \\
\cmidrule(lr){1-14}
\multirow{4}{*}{$I_{5000}$} & ORG & \textbf{\textcolor{best}{100}} & \textbf{\textcolor{worst}{99.8}} & \textbf{\textcolor{worst}{99.8}} & \textbf{\textcolor{worst}{92.5}} & \textbf{\textcolor{best}{99.8}} & \textbf{\textcolor{best}{99.8}} & \textbf{\textcolor{worst}{96.2}} & \textbf{\textcolor{best}{99.8}} & \textbf{\textcolor{best}{99.8}} & +0.2 & -7.3 & -3.6 \\
 & MS-B & 90.4 & \textbf{\textcolor{best}{93.4}} & \textbf{\textcolor{worst}{33.0}} & 96.6 & \textbf{\textcolor{worst}{93.7}} & \textbf{\textcolor{best}{100}} & \textbf{\textcolor{best}{93.5}} & \textbf{\textcolor{best}{93.5}} & \textbf{\textcolor{worst}{66.5}} & -3.0 & +2.9 & 0.0 \\
 & MS-R & 88.6 & \textbf{\textcolor{best}{92.1}} & \textbf{\textcolor{worst}{33.0}} & 96.1 & \textbf{\textcolor{worst}{92.1}} & \textbf{\textcolor{best}{100}} & \textbf{\textcolor{best}{92.3}} & 92.1 & \textbf{\textcolor{worst}{66.5}} & -3.5 & +4.0 & +0.2 \\
 & HS-R & 40.6 & \textbf{\textcolor{best}{67.5}} & \textbf{\textcolor{worst}{0.9}} & 97.6 & \textbf{\textcolor{worst}{68.4}} & \textbf{\textcolor{best}{100}} & \textbf{\textcolor{best}{69.1}} & 67.9 & \textbf{\textcolor{worst}{50.4}} & -26.9 & +29.3 & +1.2 \\
\cmidrule(lr){1-14}
\multirow{4}{*}{$I_{7500}$} & ORG & \textbf{\textcolor{best}{100}} & \textbf{\textcolor{worst}{99.8}} & \textbf{\textcolor{worst}{99.8}} & \textbf{\textcolor{worst}{93.3}} & \textbf{\textcolor{best}{99.8}} & \textbf{\textcolor{best}{99.8}} & \textbf{\textcolor{worst}{96.6}} & \textbf{\textcolor{best}{99.8}} & \textbf{\textcolor{best}{99.8}} & +0.2 & -6.5 & -3.1 \\
 & MS-B & 88.9 & \textbf{\textcolor{best}{92.4}} & \textbf{\textcolor{worst}{33.0}} & 97.5 & \textbf{\textcolor{worst}{92.0}} & \textbf{\textcolor{best}{100}} & \textbf{\textcolor{best}{93.2}} & 92.2 & \textbf{\textcolor{worst}{66.5}} & -3.4 & +5.5 & +1.0 \\
 & MS-R & 86.7 & \textbf{\textcolor{best}{90.9}} & \textbf{\textcolor{worst}{33.0}} & 95.8 & \textbf{\textcolor{worst}{90.5}} & \textbf{\textcolor{best}{100}} & \textbf{\textcolor{best}{91.2}} & 90.7 & \textbf{\textcolor{worst}{66.5}} & -4.2 & +5.2 & +0.5 \\
 & HS-R & 37.3 & \textbf{\textcolor{best}{66.1}} & \textbf{\textcolor{worst}{0.9}} & 97.7 & \textbf{\textcolor{worst}{66.6}} & \textbf{\textcolor{best}{100}} & \textbf{\textcolor{best}{67.5}} & 66.3 & \textbf{\textcolor{worst}{50.4}} & -28.8 & +31.2 & +1.2 \\
\cmidrule(lr){1-14}
\multirow{4}{*}{$I_{10000}$} & ORG & \textbf{\textcolor{best}{100}} & \textbf{\textcolor{worst}{99.8}} & \textbf{\textcolor{worst}{99.8}} & \textbf{\textcolor{worst}{93.6}} & \textbf{\textcolor{best}{99.8}} & \textbf{\textcolor{best}{99.8}} & \textbf{\textcolor{worst}{96.8}} & \textbf{\textcolor{best}{99.8}} & \textbf{\textcolor{best}{99.8}} & +0.2 & -6.1 & -3.0 \\
 & MS-B & 87.5 & \textbf{\textcolor{best}{91.7}} & \textbf{\textcolor{worst}{33.0}} & 97.4 & \textbf{\textcolor{worst}{91.9}} & \textbf{\textcolor{best}{100}} & \textbf{\textcolor{best}{92.5}} & 91.8 & \textbf{\textcolor{worst}{66.5}} & -4.1 & +5.5 & +0.7 \\
 & MS-R & 85.4 & \textbf{\textcolor{best}{90.2}} & \textbf{\textcolor{worst}{33.0}} & 96.1 & \textbf{\textcolor{worst}{90.3}} & \textbf{\textcolor{best}{100}} & \textbf{\textcolor{best}{90.7}} & 90.2 & \textbf{\textcolor{worst}{66.5}} & -4.8 & +5.8 & +0.5 \\
 & HS-R & 34.6 & \textbf{\textcolor{best}{65.5}} & \textbf{\textcolor{worst}{0.9}} & 98.4 & \textbf{\textcolor{worst}{65.8}} & \textbf{\textcolor{best}{100}} & \textbf{\textcolor{best}{66.5}} & 65.6 & \textbf{\textcolor{worst}{50.4}} & -30.9 & +32.5 & +0.8 \\
\cmidrule(lr){1-14}
\multirow{4}{*}{$I_{12500}$} & ORG & \textbf{\textcolor{best}{100}} & \textbf{\textcolor{worst}{99.8}} & \textbf{\textcolor{worst}{99.8}} & \textbf{\textcolor{worst}{93.9}} & \textbf{\textcolor{best}{99.8}} & \textbf{\textcolor{best}{99.8}} & \textbf{\textcolor{worst}{96.9}} & \textbf{\textcolor{best}{99.8}} & \textbf{\textcolor{best}{99.8}} & +0.2 & -5.8 & -2.8 \\
 & MS-B & 86.4 & \textbf{\textcolor{best}{91.1}} & \textbf{\textcolor{worst}{33.0}} & 97.4 & \textbf{\textcolor{worst}{91.2}} & \textbf{\textcolor{best}{100}} & \textbf{\textcolor{best}{91.9}} & 91.2 & \textbf{\textcolor{worst}{66.5}} & -4.7 & +6.1 & +0.7 \\
 & MS-R & 84.3 & \textbf{\textcolor{best}{89.5}} & \textbf{\textcolor{worst}{33.0}} & 96.5 & \textbf{\textcolor{worst}{89.4}} & \textbf{\textcolor{best}{100}} & \textbf{\textcolor{best}{90.4}} & 89.4 & \textbf{\textcolor{worst}{66.5}} & -5.2 & +7.1 & +0.9 \\
 & HS-R & 33.2 & \textbf{\textcolor{best}{64.1}} & \textbf{\textcolor{worst}{0.9}} & 98.4 & \textbf{\textcolor{worst}{65.4}} & \textbf{\textcolor{best}{100}} & \textbf{\textcolor{best}{65.8}} & 64.8 & \textbf{\textcolor{worst}{50.4}} & -30.9 & +33.0 & +1.0 \\
\bottomrule
\end{tabular}%
}
\end{table*}

\bibliographystyle{IEEEtran}
\bibliography{bibliography}

\end{document}